\documentclass[conference]{IEEEtran}
\IEEEoverridecommandlockouts
\usepackage[square,sort,comma,numbers]{natbib}
\setcitestyle{citesep={,}}
\usepackage{amsmath,amssymb,amsfonts}
\usepackage{adjustbox}
\usepackage{algpseudocode}
\usepackage{graphicx}
\usepackage{multirow}
\usepackage{physics}
\usepackage[algo2e]{algorithm2e} 
\usepackage[utf8]{inputenc}
\usepackage{graphicx,wrapfig,lipsum}
\usepackage{listings}
\usepackage{float}
\usepackage{amsfonts}
\usepackage{algorithm}
\usepackage{algcompatible}
\usepackage{natbib}
\usepackage{color}   
\usepackage{hyperref}
\hypersetup{
    colorlinks=false, 
    linktoc=all,     
    linkcolor=blue,  
}
\usepackage{caption}
\usepackage{subfig}

\usepackage{stmaryrd} 
\usepackage{amsmath}
\usepackage{amssymb}
\usepackage{amsfonts}
\usepackage{color}
\usepackage{textcomp}
\setcitestyle{square}

\usepackage{xcolor}
\def\BibTeX{{\rm B\kern-.05em{\sc i\kern-.025em b}\kern-.08em
    T\kern-.1667em\lower.7ex\hbox{E}\kern-.125emX}}
\begin{document}
\title{Knowledge-based Deep Learning for Modeling Chaotic Systems}

 \author{\IEEEauthorblockN{ Zakaria Elabid}
 \IEEEauthorblockA{\textit{Sorbonne Center for Artificial Intelligence} \\
 \textit{Sorbonne University Abu Dhabi}\\
 Abu Dhabi, UAE\\
 Zakaria.Alabid@sorbonne.ae}
 \and
 \IEEEauthorblockN{Tanujit Chakraborty}
 \IEEEauthorblockA{\textit{Dept. of Science and Engineering} \\
 \textit{Sorbonne University Abu Dhabi}\\
 Abu Dhabi, UAE\\
 tanujit.chakraborty@sorbonne.ae}
 \and
 \IEEEauthorblockN{Abdenour Hadid}
 \IEEEauthorblockA{\textit{Sorbonne Center for Artificial Intelligence} \\
 \textit{Sorbonne University Abu Dhabi}\\
 Abu Dhabi, UAE\\
 Abdenour.Hadid@sorbonne.ae}
 }

\maketitle

\begin{abstract}
Deep Learning has received increased attention due to its unbeatable success in many fields, such as computer vision, natural language processing, recommendation systems, and most recently in simulating multiphysics problems and predicting nonlinear dynamical systems. However, modeling and forecasting the dynamics of chaotic systems remains an open research problem since training deep learning models requires big data, which is not always available in many cases. Such deep learners can be trained from additional information obtained from simulated results and by enforcing the physical laws of the chaotic systems. This paper considers extreme events and their dynamics and proposes elegant models based on deep neural networks, called knowledge-based deep learning (KDL). Our proposed KDL can learn the complex patterns governing chaotic systems by jointly training on real and simulated data directly from the dynamics and their differential equations. This knowledge is transferred to model and forecast real-world chaotic events exhibiting extreme behavior. We validate the efficiency of our model by assessing it on three real-world benchmark datasets: El Niño sea surface temperature, San Juan Dengue viral infection, and Bjørnøya daily precipitation, all governed by extreme events' dynamics. Using prior knowledge of extreme events and physics-based loss functions to lead the neural network learning, we ensure physically consistent, generalizable, and accurate forecasting, even in a small data regime. 
\end{abstract}

\begin{IEEEkeywords}
Chaotic systems, long short-term memory, deep learning, extreme event modeling. 
\end{IEEEkeywords}

\section{Introduction}
Centuries-old efforts to comprehend and forecast the dynamics of chaotic systems have spurred developments in large-scale simulations, dimensionality reduction techniques, and a multitude of forecasting techniques \cite{vlachas2018data, klos2020dynamical}. Chaotic systems describe deterministic dynamical systems that exhibit random behavior after a while and cannot be predicted sometimes~\cite{tabor1989chaos, wiggins2003introduction, lakshmanan2012nonlinear}. These systems can be modeled with physical laws to express their temporal and spatial dependencies \cite{PINN, BPINN}. Some examples of these chaotic systems are: daily weather forecasts, viral diseases (e.g., Covid-19), traffic jams, double pendulum \cite{pendulum}, etc. Modeling chaotic systems has always been an important study in various fields, from geophysics \cite{Geophysics}, atmospheric physics \cite{weather}, electronics \cite{electronics}, and biochemistry \cite{biochemistry} among many others. Various dynamical systems evolved in order to understand natural phenomena and their unpredictable behavior and how a perturbation on initial conditions in real-world problems can randomly change their outcome \cite{klos2020dynamical}. The goals of understanding and predicting chaotic systems using physical laws and data-driven approaches have complemented each other for the last few decades. In recent years, we have observed a convergence of these approaches towards modern data-driven methodologies due to advances in computing power, algorithmic innovations, and the ample availability of data. A major beneficiary of this convergence are physics-informed machine learning methods \cite{PINN,karniadakis2021physics, PIML}, dynamical learning with deep neural networks \cite{Ensemble} and representation learning tools to study continuous dynamical systems \cite{klos2020dynamical} and molecular properties \cite{satorras2021n}. All these methods aim to provide precise short-term predictions while capturing the long-term trajectories of the systems. Successful deep learning-based forecasting methods (such as long short-term memory networks (LSTM) \cite{lstm}, convolutional neural networks (CNN) \cite{CNN}, feed-forward neural networks (FFNN) \cite{FFNN}, reservoir computing (RC) \cite{Reservoir}) can capture the evolution of time series in chaotic systems but usually do not use physical laws in their architecture.

Chaotic systems are widely used to describe extreme events, such as natural catastrophes and weather dynamics, which are highly sensitive to fluctuations \cite{Chaos}. While the weather can be predicted for the near future, some random factors make it impossible to forecast as time goes on. Currently, chaotic systems are modeled through dynamical systems \cite{klos2020dynamical} - a differential equation of the function describing the spatio-temporal dependency between the systems' variables. For example, sea surface temperature is governed by mass and energy conservation laws: Navier–Stokes equations and heat transport equation \cite{transport1}. While physics-driven approaches manage to predict chaotic systems in the short-term, they fail to accurately forecast real-world phenomena when noise or perturbation is present, which is always the case in chaotic systems \cite{unpredictability2}. In addition, due to the complexity of the dynamics, some chaotic systems cannot be solved analytically and instead are modeled through direct numerical simulation, decreasing the reliability of the prediction and increasing the uncertainty of the analysis of the physical processes \cite{uncertainty}. On the other hand, data-driven approaches to model and forecast chaotic systems using deep neural networks require large amounts of data to train the models. They are difficult to generalize and mostly fail to emulate long-time dependencies when the system's behavior is random. Multiple studies focused on modeling chaotic systems using deep learning, especially when experimental data is small, by integrating prior knowledge \cite{Ensemble, vlachas2018data, klos2020dynamical, PIML}.

Incorporating physical knowledge to model real-life phenomena using deep learning models is a promising field of study, potentially capable of overcoming the everlasting challenges of modeling chaotic systems. Physics-informed machine learning (PIML), a new variant of machine learning models, embeds prior physical knowledge in data-driven algorithms through the use of differential equations \cite{PINN, PGNN, PIML}. This method is aptly used when a large amount of temporal data is unavailable for making accurate and reliable forecasts of chaotic systems containing extreme events. Instead of entirely relying on the data to capture the complex underlying phenomena, it incorporates physical knowledge into data-driven deep learning systems, see for detailed surveys \cite{karniadakis2021physics, PIML} (see also references therein). In recent seminal work, physics-informed neural networks (PINN) have been introduced as a function approximator \cite{PINN}. By providing several collocation points, the initial and boundary conditions, PINN manages to provide an accurate solution to the governing partial derivative equations (PDE) through a regularization term in the loss function. PINN can also be used to solve inverse problems when coefficients of the PDE, initial or boundary conditions are rebuilt and the solution of the governing PDE is partially known. Among many applications, PINN was used to simulate the flow in an expresso cup \cite{expresso}, modeling 3D temperature data using physical laws (mass, momentum, and energy conservation) in a fully connected network architecture. To predict the effect of cloud processes on climate, authors enforced physical constraints to machine precision (hard-enforcement) by replacing layers of a neural network with the help of conservation laws \cite{Machineprecision}. Another similar architecture, namely physics-guided neural networks (PGNN), is used to model lake temperature as a spatio-temporal forecasting problem in which LSTM networks build the model using the dynamics of sea surface temperature \cite{PGNN}. More precisely, the conservation laws enforce the monotonicity of density with depth and incorporate the dynamics by hard encoding them on a machine precision in a PGNN architecture for a small data regime.

However, the models described above fall short when it comes to modeling chaotic systems. PINN can only be used to emulate data, i.e., it numerically simulates the solution to PDEs governing the dynamics of the system. Furthermore, PINN requires collocation points, boundary, and initial conditions as input. Therefore, these models (e.g., PINN) only work on simulated data and cannot be directly generalized to real-world problems in modeling chaotic systems. On the other hand, physics-guided architectures can forecast lake surface temperature \cite{PGNN} since the equation governing the dynamics of temperature and density is of order zero and does not require their spatio-temporal derivatives. Therefore, such a model cannot be used to emulate more complex dynamics governed by non-linear partial differential equations of higher order. 

Motivated by the shortcomings of the aforementioned models, we aim to design a physically consistent and generalizable deep learning model for modeling chaotic systems that seamlessly integrate real data and simulated data directly from the dynamics and their differential equations. Hence, we introduce 
the knowledge-based deep learning (KDL) approach. Our proposed KDL approach learns the extreme events' dynamics and temporal patterns for the observed chaotic systems. Our main contributions in this paper are described as follows:
\begin{itemize}
\item We propose a new framework called KDL (knowledge-based deep learning), in which the network is trained on synthetic data simulated from the dynamics and transfers the knowledge to real-world data governed by the same physical laws. Incorporating the dynamics through a regularization term in the loss function corresponding to the physical law governing the system makes the deep learning framework suitable for small data regime problems in chaotic systems.
\item KDL algorithm computes discrete temporal derivatives and includes them during the backpropagation in the loss function of the LSTM network.
\item We evaluate our proposed approach on several time series problems containing extreme events and demonstrate superior performance compared to several state-of-the-art deep learning models.
\end{itemize}





\section{Preliminaries}
This section discusses some necessary mathematical preliminaries on the Liénard-type system, followed by a summarized introduction on the long-short term memory network and transfer learning used in building our proposed KDL approach. 

\subsection{Liénard-type System}
Nonlinear oscillator systems are omnipresent in physics and are used to model numerous physical phenomena ranging from atmospheric physics, condensed matter, and nonlinear optics to electronics, plasma physics, biophysics, and evolutionary biology, among many others \cite{tabor1989chaos, wiggins2003introduction, lakshmanan2012nonlinear}. For example, Liénard-type nonlinear oscillator systems are very present in chaos theory and extreme events studies \cite{chandrasekar2005unusual}. A Liénard equation is a second-order differential equation in mathematics specifically used in the study of dynamical systems. For any two functions $f$ and $g$ on $\mathbb{R}$, the second-order differential equation of the following form is called a Linéard equation: 
$$\frac{d^{2} x}{d t^{2}}+f(x) \frac{d x}{d t}+g(x)=0.$$
Liénard-type system has a unique limit cycle, i.e., when time is close to infinity, at least one trajectory spirals into a closed trajectory \cite{sugie2022qualitative}. The Liénard-type oscillator with a periodic forcing, exhibits extreme events for an admissible set of parameter values, as
\begin{align}
\frac{dx}{dt} & =y \nonumber \\ 
\frac{dy}{dt} & =-\alpha x y-\gamma x-\beta x^{3}+f \sin (\omega t),
\label{eq:lienard}
\end{align}
where $\alpha$ and $\beta$ are the nonlinear dampings and the strength of nonlinearity, respectively, and $ \gamma$ is related to the internal frequency of the autonomous system. $f$ and $\omega$ are the amplitude and the frequency of an external sinusoidal signal, respectively~\cite{Lienard-extreme}. An extremely large amplitude of intermittent spikings in a dynamical variable of a periodically forced Liénard-type oscillator can be observed, and thus, it can be characterized as extreme events, which are rare but recurrent and larger in amplitude than a threshold.


\subsection{Long-Short Term Memory and Transfer Learning}
Long-Short Term Memory networks (LSTM) are an evolution of recurrent neural networks (RNN) conceived to overcome the stability and speed issues in classical RNN \cite{long-short}. They are omnipresent in speech recognition problems, machine translation, image captioning, and time series forecasting, among many other applications. They are also widely used to model chaotic systems due to their ability to memorize long-term dependencies in the training data, which helps deal with the problem of vanishing gradients and the long-term randomness~\cite{LSTMchaos}. LSTM regulates the flow of sequential inputs through the memory cells of the network by employing three main gates: \textit{input gate}, \textit{output gate}, and a \textit{forget gate} that decide which bits of long-term dependencies are relevant when given previous hidden state and current state input data. However, the time and computational cost required to train this sequential neural network become an overarching problem. Among many solutions to this problem, using pre-trained models is a popular approach to decrease time, increase performance, and improve the model's generalization. One such approach is Transfer Learning (TL), which focuses on storing knowledge gained from solving one problem and applying it to different but related problems \cite{transfer}. We used TL methodology in our application to transfer the knowledge from a synthetic dataset simulated from a Liénard-type differential equation to real-world datasets, exhibiting extreme events and following the same dynamics.



\section{Proposed Method: Knowledge based deep learning (KDL) }
We consider a memristor Liénard system with external harmonic perturbation from Eqn. \ref{eq:lienard}
where the parameters $f, \alpha, \gamma, \omega, \beta$ are known. 
We generate our synthetic dataset by simulating data points from the nonlinear differential equations using the Runge-Kutta method \cite{Runge-kutta}. We train an LSTM network on the simulated time series while enforcing the physical law
on the network as a regularization term. The network learns complex patterns from historical values and the physical distribution of the target values. In the standard PINN model, the time index is considered an input, and the output is the solution of the differential equation. Computing the regularization term amounts to differentiating the multilayered perceptron (MLP) network and computing the time derivatives using auto-differentiation \cite{autodiff}. However, time series are discrete observations and there is not a substantial mathematical equation governing the observed variables in time, and subsequently, it is difficult to compute the derivatives in time. To overcome this, we compute discrete derivatives of the time series. For an observed time series $x(t)$ indexed over time $t$, the discrete-time derivative of order one can be written as:
$$\frac{dx}{dt} = \frac{x(t+\delta t)-x(t)}{\delta t},$$ 
where $\delta t$ is the lag. For real-world time series datasets, time is recorded as a timestamp (date) in chronological order. Thus, we choose $\delta t = 1$ in our work. To adjust the loss of the network, the standard PINN model backpropagates the time derivatives since the mapping function used in MLP is differentiable.

For instance, if $x = g(t, \theta=(w,b))$, where $g$ is the neural network function of the MLP, then $y = \frac{dx}{dt} = \frac{dg(t, \theta=(w,b))}{dt}$. The backpropagation writes:
\begin{equation*} 
\begin{split}
\frac{dL_{phy}}{d\theta} 
        & = \frac{dL_{phy}}{dy}\frac{dy}{d\theta} 
          = \frac{dL_{phy}}{dy}\frac{d^2x}{dtd\theta} \\
        & = \frac{dL_{phy}}{dy}\frac{d^2x}{dtd\theta}  
          = \frac{dL_{phy}}{dy}\frac{d^2g(t, \theta)}{dt d\theta}, \\
\end{split}
\end{equation*}
where $L_{phy}$ is the physical loss corresponding to the regularization term and $\theta = (w,b)$ represents the weights and biases of the network. In our case, we computed the time derivatives during the pre-processing step. 

Our network forecasts $x(t)$, $\frac{dx}{dt}$ and $\frac{d^2x}{dt^2}$ in time from their historical data. The physics is enforced on the system through regularization of the output, i.e., the predicted values of time series $x(t)$ and its derivatives have to satisfy the Liénard equation as in Eqn  \ref{eq:lienard}.
After pre-training, we transfer this knowledge to real-life data following the same dynamics. The importance of transferring synthetic data knowledge to real data is further experimentally explored in sections \ref{experiments} and \ref{lastx}. In the case when harmonic forcing (second-term of the differential equation) is not known, we assume it can be approximated by the $1^{st}$ term of the differential equation for the real data distribution. 


\begin{equation}
\begin{split}
L_{phy}=RMSE&(\frac{d^2y_{pred}}{dt^2} + \alpha y_{pred}\frac{dy_{pred}}{dt} + \gamma y_{pred}
    +\beta y_{pred}^3,\\
    &fsin(\omega t))
\end{split}
    \label{synth}
\end{equation}

\begin{figure*}
    \centering
    \includegraphics[width = 0.9\textwidth]{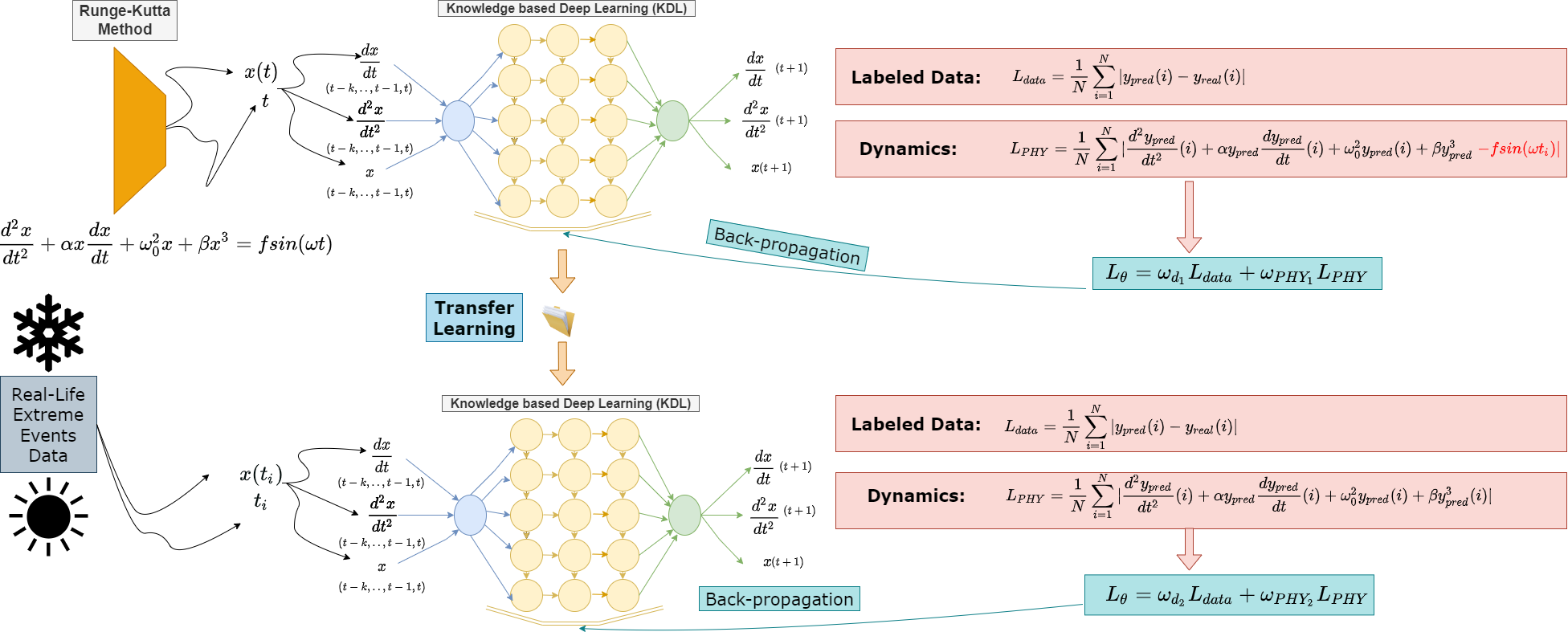}
    \caption{Proposed knowledge-based deep learning architecture. Differential equation is simulated into a time series $t,x(t)$. The first and second-order derivatives of $x$ are computed and fed along with $x$ to our model that forecasts the next instances through a backpropagation mechanism of the losses: 1) Data loss 2) Physics loss: Predicted data should satisfy the dynamics. The same process is used on real-world datasets with the addition of transferred knowledge from the pre-training.}
    \label{fig:Liénard}
\end{figure*}

In our case, we minimize the gap between the $1^{st}$ term of the differential equation of the real and the predicted data as follows:
\begin{equation}
    \begin{split}
L_{phy}=RMSE & (\frac{d^2y_{real}}{dt^2} + \alpha y_{real}\frac{dy_{real}}{dt} + \gamma y_{real}
    +\beta y_{real}^3 , \\
    &\frac{d^2y_{pred}}{dt^2} + \alpha y_{pred}\frac{dy_{pred}}{dt} + \gamma y_{pred}
    +\beta y_{pred}^3),
    \end{split}
    \label{real}
\end{equation} 
where RMSE is the root mean squared error. A schematic diagram showing the various steps of the KDL approach is presented in Figure \ref{fig:Liénard}. An algorithmic version of the proposed framework is also given in Algorithm \ref{KDL_algo}.

\begin{algorithm}
\caption{\textsc{Knowledge-Based Deep Learning (KDL)}}\label{KDL_algo}
\KwData{X = $x(t-k),\ldots,x(t-1), x(t)$,$\frac{dx}{dt}(t-k),\ldots,\frac{dx}{dt}(t)$, $\frac{d^2x}{dt^2}(t-k),\ldots,\frac{d^2x}{dt^2}(t)$}

\KwResult{Y = $x(t+1)$, $\frac{dx}{dt}(t+1)$, $\frac{d^2x}{dt^2}(t+1)$}

Initialize network parameters: (weights and biases)\;

\eIf{Pre-training}{
\While{$epoch<max\_epochs$ \textbf{or} Termination Condition}{
Compute Y = KDL(X)\;

Compute the loss $L_{data} = RMSE(Y_{predicted}, Y_{real})$\;

Compute the loss $L_{phy}$ from Equation \eqref{synth}\;

Backpropagate $L_{data} + \lambda_1 L_{phy}$ (using ADAM optimizer) and $\lambda_1$ is a tuning parameter chosen by cross-validation\;
 
Save Network State (weights, biases)\;
  
}
}{
\While{$epoch<max\_epochs$ \textbf{or} Termination Condition} {
Load pretrained network state (weights, biases)\;

Compute Y = KDL(X)\;
    
Compute the loss $L_{data} = RMSE(Y_{predicted}, Y_{real})$\;
   
Compute the loss $L_{phy}$ from Equation \eqref{real}\;

 Backpropagate $L_{data} + \lambda_2 L_{phy}$ (using ADAM optimizer) and $\lambda_2$ is a tuning parameter chosen by cross-validation\;
}}

\end{algorithm}

\section{Experimental Setup}\label{experiments}
We evaluate KDL on three datasets containing extreme events to validate our proposed model. 
Extreme events can be modelled as a Liénard system with external forcing from Equation \ref{eq:lienard}, and for a set value of parameters \cite{Lienard-extreme}: $f = 0.2$, $\alpha = 0.45$, $\gamma = -0.5$, $\omega = 0.642$, $\beta = 0.5.$

\subsection{Real-world Datasets}

\subsubsection{El Niño Dataset}
El Niño dataset, also known as ENSO, is a periodic fluctuation of sea surface temperature (SST) across the Pacific Ocean \cite{Ensemble}. El Niño dataset contains 1634 weekly observations of SST in El Nino region 1 across the Pacific Ocean, from January 3, 1990 to April 21, 2021.

\subsubsection{San Juan Dengue infections}
San Juan dataset \cite{chakraborty2019forecasting} is a univariate time series data portraying the evolution of cases of Dengue infection. This infection exhibits extreme events due to the randomness of the transmission. Dengue infection is a mosquito-borne viral disease. The dataset contains 1197 observations from week 17 of 1990 to week 16 of 2013. We consider only the univariate time series with the number of cases across this time period.

\subsubsection{Bjørnøya Precipitation} 
This is an extreme event dataset taken from the Norwegian Climate Service Center \cite{norsk} observation of rainfall in the Bjørnøya region. The dataset consists of 15320 daily observations from June 16, 1980 to June 16, 2022.

\subsection{Performance Measures}
To evaluate our models, we considered three commonly used metrics as given below \cite{Statisticaltests}. The lower their value is, the better the forecasting model. We have used the following measures in this study: (a) Root Mean Squared Error (RMSE) \cite{Statisticaltests}; (b) Mean Absolute Error (MAE); and (c) Physical inconsistency (PIC) \cite{PGNN}.

    
Among the three performance measures, physical inconsistency estimates the physical meaning of a variable. It is computed by comparing the dynamics of real data to the predicted data. This metric is relevant in problems revolving around physics. For example, to predict temperature T at a time point using two different models knowing that $T_{real}= 25$, if the model 1 predicted $T_{pred}=24.9$ and model 2 predicted $T_{pred}=26.1$, model 1 outperforms model 2. Nonetheless, under the hypothesis that temperature $T\geq 25$ (from the dynamics of the system), then model 2 is more relevant, even when the average error is higher.

\subsection{Implementation of the proposed KDL framework}
Data is generated from the differential equation using the Runge-Kutta method and ODE-4 Solver in Matlab. For the rest, we use Keras's backend. We train our model on data simulated from the Liénard-type differential equation \cite{Ensemble}. The result is a time series $t,x(t)$. We compute the first derivative by shifting x by a step of one and subtracting x from it i.e., $\frac{dx}{dt}=x_{t+1}-x_t$. The 2nd derivative is computed similarly from $\frac{dx}{dt}$. We transform our time series into supervised learning $(X,Y)$ where $X = (x(t-p),...x(t-1))$ is the input and \mbox{Y = $(x(t),\frac{dx}{dt}(t),\frac{d^2x}{dt^2}(t))$} is the output and $p$ is the lookback (in our case $p$=10).
We feed data to our network composed of three LSTM layers with 50 neurons each and a Dense layer with 3 neurons (3 outputs).
The feedback mechanism has the standard data loss: $RMSE(y_{predicted},y_{real})$ and a physical loss ensuring the differential equation (as in Equation \ref{eq:lienard}) is satisfied.
The total loss is expressed as $L_{total}=L_{data}+0.2L_{physical}$.

We save the network state after the training.
For real data, we follow the same procedure by loading the state of the network that we saved from the pre-training when we initialized our model before training it.



\begin{table*}
    \begin{center}
      \caption{Preliminary data analysis}
      \label{tab:Data}
      \renewcommand{\arraystretch}{2}
       \begin{adjustbox}{width=\textwidth}
      \begin{tabular}{|l|c|c|c|c|c|c|} 
      \hline
        \textbf{Data Name} & \textbf{Frequency} & \textbf{Observations} & \textbf{Start} &\textbf{End} & \textbf{Min-Max Value} & \textbf{Behaviour}  \\
        \hline
          El Niño Dataset \cite{Ensemble} & Weekly & 1634 &
January, $3^{rd}$ 1999 & April, $21^{st}$  2021 & 18.3 - 29.2 & Stationary, Non-linear, Chaotic, Long-term dependent\\
        \hline
          San Juan Dengue cases \cite{chakraborty2019forecasting} & Weekly & 1197 &April, $23^{rd}$ 1990 & April, $21^{st}$ 2013 & 0 - 461 & Non-stationary, Non-linear, Chaotic, Long-term dependent\\
        \hline
          Bjørnøya Precipitation \cite{norsk} & Daily & 15320 &June,
$16^{th}$ 1980 &June, $16^{th}$ 2022 & 0.0 - 42.1 & Non-stationary, Non-linear, Chaotic, Long-term dependent\\ 
        \hline
      \end{tabular}
       \end{adjustbox}
      \renewcommand{\arraystretch}{1}
    \end{center}
 \end{table*} 

\vspace{-1mm}
\section{Experimental Results} \label{lastx}
\vspace{-1mm}
\subsection{Preliminary data analysis}
Table \ref{tab:Data} displays the statistical specifications of our datasets including stationarity (using Kwiatkowski–Phillips–Schmidt–Shi test), nonlinearity (using Teräsvirta’s neural network test), long-term dependency (using Hurst-Exponent), and evidence of chaos (using Lyapunov Exponent test) as described in \cite{Statisticaltests}. Our datasets exhibit strong nonlinearity, long-term dependency, chaotic behavior, and non-stationarity (except for the El Niño dataset's stationary structure).

\subsection{Baseline models}
For comparative analysis, we considered the following competitive forecasting models: Long-Short Term Memory (LSTM) \cite{lstm}; Feed-Forward Neural Networks \cite{FFNN}; Convolutional Neural Networks (1D CNN) \cite{CNN}, and Reservoir Computing (ESN) \cite{Reservoir}. 
In the experiments, we fit an LSTM with 3 hidden layers and 50 neurons, an RC with a reservoir size of 500, an FFNN with one hidden layer and 50 neurons, a 1D CNN with 64 filters and a kernel size of 2, and a timestep of 3 followed by a MaxPooling and Flatten layers to a fully connected layer with 50 neurons. Our network has the same architecture as the LSTM used for comparison, with the addition of physical regularization and prior knowledge from pre-training. All the methods are extensively evaluated on different train/test splits scenarios to assess the performance of the different methods under different data sizes.

\subsection{Results}

\begin{table*}[t]
\begin{center}
\captionsetup{justification=centering}
\caption{Comparison of baseline models: LSTM, FFNN, CNN, ESN with our model KDL on three real-world datasets: El Niño, San Juan dengue, and Bjørnøya precipitation on different training sizes: 20\%, 40\%, 60\%, and 80\%.}
 \begin{adjustbox}{width=\textwidth}
\begin{tabular}{|c| c | c | c | c | c |c |c |c |c |c |c |c |c |c |c |c |}
\hline
\multirow{2}{4em}{\textbf{Dataset}}&\textbf{Network/Loss} & \multicolumn{3}{ c |}{\textbf{LSTM \cite{lstm}}} & \multicolumn{3}{ c |}{\textbf{FFNN \cite{FFNN}}} & \multicolumn{3}{ c |}{\textbf{CNN \cite{CNN}}}& \multicolumn{3}{ c |}{\textbf{Reservoir Computing \cite{Reservoir}}}& \multicolumn{3}{ c |}{{\textcolor{blue}{KDL (proposed)}}} \\
\cline{2-17}
&\textbf{Train : Test split}& RMSE & MAE & PIC & RMSE & MAE & PIC & RMSE & MAE & PIC & RMSE & MAE & PIC & RMSE & MAE & PIC  \\
\hline
\multirow{4}{4em}{El Niño}&20\% : 80\%& 0.433 & 0.342 & 283.340 & 2.444 & 2.108 & 1713.237 & 0.523 & 0.413 & 339.859 &1.399&0.925&733.396&\textbf{0.428}&\textbf{0.338}&\textbf{276.654}\\ 
&40\% : 60\% & 0.418 &  \textbf{0.327} &  \textbf{268.309} & 1.948 &  1.677 &  1349.331& 0.438 & 0.378 & 313.898&2.889&1.685&1176.900&\textbf{0.416}&\textbf{0.327}&268.777\\ 
&60\% : 40\%&  0.433 & 0.343 &  285.274 &  2.068 & 1.780 &  1456.752& 0.452 & 0.350 & 293.451&0.694&0.503&395.430&\textbf{0.431}&\textbf{0.341}&\textbf{283.476}\\ 
&80\% : 20\%& \textbf{0.427} & \textbf{0.333} & \textbf{280.754} & 2.157 & 1.846 & 1540.997& 0.490 & 0.394 & 336.259&0.562&0.439&359.607&\textbf{0.427}&0.336&283.362\\ 
\hline
\multirow{4}{4em}{San Juan Dengue}&
20\% : 80\%& 15.081 & 9.214 & \textbf{130027} & 38.995 & 33.393 & 252054.109 & 17.037 & 9.536 & 156565.047 &20.722&13.763&182934.594&\textbf{14.333}&\textbf{8.537}&133578\\
&40\% : 60\%  & 11.402 &  7.456 &  \textbf{89123} & 34.753 &  25.631 &  231226.234& 14.152 & 9.218 & 127899.242 &12.031&8.140&89863.445&\textbf{11.328}&\textbf{7.352}&90754\\
&60\% : 40\%&  13.131 & 8.565 &  136547 &  38.039 & 24.461 &  330925.906& 17.266 & 10.259 & 184430.344&13.289&8.779&129067.086&\textbf{12.894}&\textbf{8.358}&\textbf{128159}\\ 
&80\% : 20\% & 15.562 &10.803&235978 & 49.334 & 30.185 & 609710.938& 19.739 & 12.755 & 278096.344&15.420&10.762&235443.344&\textbf{15.416}&\textbf{10.721}&\textbf{231362}\\ 
\hline
\multirow{4}{4em}{Bjørnøya Rainfall}&
20\% : 80\%& 2.474 & 1.328 & 51.775 & 2.506 & 1.348 & 51.608 & 3.179 & 1.712 & 88.626 &2.474&1.351&\textbf{50.932}&\textbf{2.473}&\textbf{1.322}&51.760\\ 
&40\% : 60\% & 2.507&  \textbf{1.380} &  53.719 & 2.543 &  1.448 &  53.562& 3.071 & 1.687 & 69.999 &2.510&1.383&\textbf{52.821}&\textbf{2.506}&1.403&53.705\\ 
&60\% : 40\% &  2.428 & 1.418 &  47.696 &  2.457 & 1.456 &  47.450& 2.923 & 1.527 & 65.802&2.428&1.389&\textbf{46.876}&\textbf{2.427}&\textbf{1.383}&47.671\\ 
&80\% : 20\%& \textbf{2.561} &1.369&63.420 & 2.612 & 1.498 & 63.328& 3.167 & 1.583 & 120.577&2.563&1.930&\textbf{62.666}&2.562&\textbf{1.368}&63.485\\ 
\hline
\end{tabular} \label{T:equipos}
  \end{adjustbox}
\end{center}
\end{table*}


\subsubsection{El Niño dataset}
On the El Niño dataset, we can see that the RMSE of our method is lower than that of 4 other forecasting models. The lower the RMSE, the better we consider our prediction. As for MAE, our network gets the highest score for training/test sizes up to 60\%, being a close second with 80\% training and 20\% test (0.336 vs. 0.333 MAE). Our proposed model has lower physical inconsistency than all other prediction models, especially when trained on low data sizes. For higher data sizes, our network is close to LSTM as classical approaches can capture complex dynamics when fed with enough data points.
\subsubsection{San Juan Dengue}
Our proposed network manages to get better physical inconsistency with higher training sizes on San Juan Dengue datasets. This could be explained by the fact that this system exhibits extreme events globally but not locally. Therefore, we removed the prior training (on simulated data) and trained the network solely on this dataset. Having prior knowledge of different dynamics further degrades the accuracy of the model.

\subsubsection{Bjørnøya daily precipitation}
The data exhibits chaos when the total number of observations is high enough (15k observations in this case as opposed to ~1.5k for other datasets). Traditional deep learning methods can capture the complex patterns and dynamics in data when fed with enough data points which can be further validated when we look at the physical inconsistency values where our network falls short compared to Reservoir Computing. Our network scores best RMSE for training sizes up to 60\% and 2nd place when the training size is 80\%. For MAE, we get 1st rank when training sizes are respectively 20\%,60\% and 80\%.

\subsection{Comparison with baselines}
\begin{figure}[h!]
    \centering
    \includegraphics[width = 0.35\textwidth]{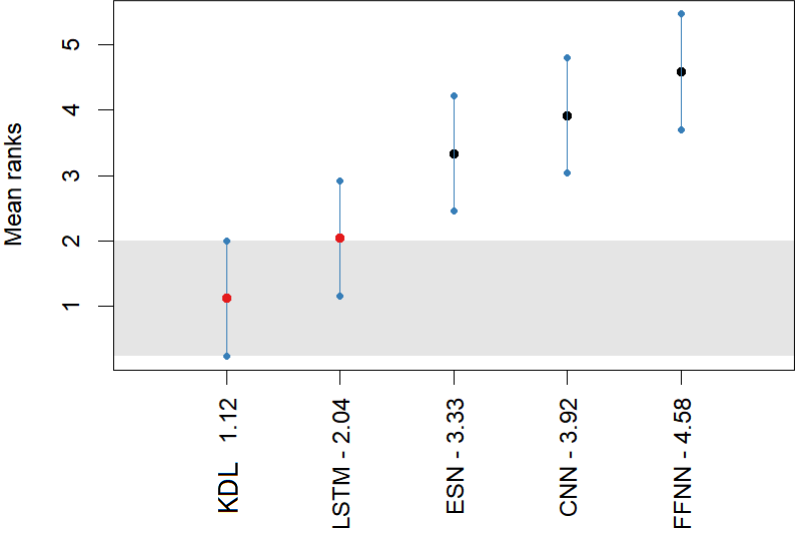}
    \caption{Comparison with baseline models based on RMSE. KDL scores an average rank of 1.12, the highest rank among baseline models, followed by LSTM, ESN, CNN, and FFNN.}
    \label{fig:my_label1}
\end{figure}

\begin{figure}[h!]
    \centering
    \includegraphics[width = 0.35\textwidth]{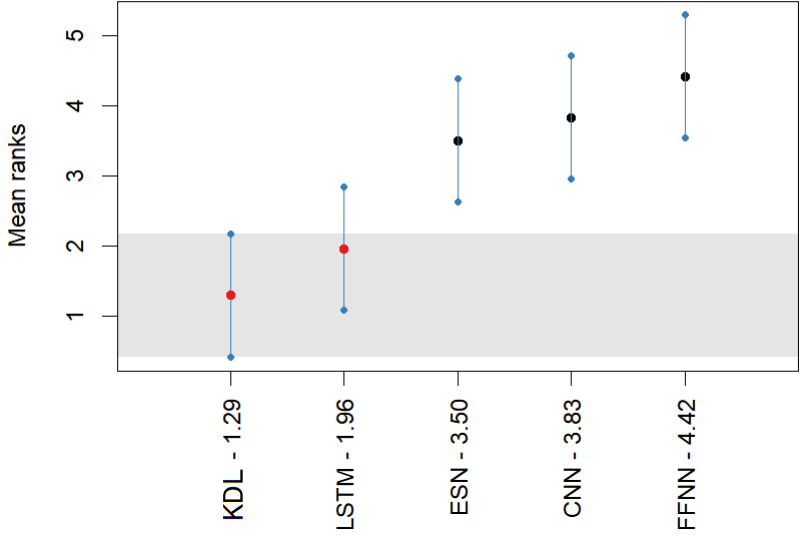}
    \caption{Comparison with baseline models based on MAE. KDL scores an average rank of 1.29, the highest rank among baseline models, followed by LSTM, ESN, CNN, and FFNN.}
    \label{fig:my_label2}
\end{figure}

Table \ref{T:equipos} depicts the experimental results and comparison with several baseline models. We further apply multiple comparisons with the best (MCB) test to check the statistical significance of the comparative performance \cite{panja2022epicasting}. Finally, based on the results obtained in the previous section, we plot the average ranking of the models on all datasets and all training: test sizes based on MAE and RMSE. From figures \ref{fig:my_label1} and \ref{fig:my_label2}, we can conclude that our proposed KDL method outperforms all other models in terms of RMSE and MAE. All the code and material are made available at \url{https://github.com/Zelabid/KdL} to support the principle of reproducible research.

\vspace{-1mm}
\section{Conclusion and Discussion}
In this paper, we proposed a knowledge-based deep learning framework combining physical and prior transferable knowledge from synthetic and real data to forecast extreme events. The proposed model takes a hybrid approach, learning partially from the time series, their physics, and the knowledge gained in prior training. We built our model under the hypothesis that chaotic systems exhibit extreme events and can be modeled using a Liénard-type system. This assumption holds strongly as our network ranks first among all other forecasting models considered in the comparison. It is interesting to consider the proposed KDL approach with multi-step forecasting for future work. Generalizing the framework to other relevant forecasting problems can also be regarded as future scope of this study.


 \section*{Acknowledgment}
The support of TotalEnergies is fully acknowledged. Zakaria ELabid (PhD Student) and Abdenour Hadid  (Professor,  Industry Chair at SCAI Center of Abu Dhabi) are funded by TotalEnergies collaboration agreement with Sorbonne University Abu Dhabi.

\bibliographystyle{IEEEtran}
\bibliography{ref}

\end{document}